\documentclass[letterpaper, 10 pt, journal, twoside]{ieeetran}

\IEEEoverridecommandlockouts                              

\usepackage{graphicx, soul, color, bm, bbm}
\usepackage{tikz}
\usepackage{amsmath, amssymb, mathrsfs, amsopn}
\usepackage{multirow}
\usepackage{epstopdf, epsfig}
 \usepackage{subcaption, caption, hyperref, xcolor}
\usepackage{float}
\usepackage{multicol, pdfpages}
\usepackage[smartEllipses]{markdown}

\usepackage{verbatim}
\definecolor{hlColor}{HTML}{1434A4}
\sethlcolor{hlColor}
\definecolor{todoColor}{HTML}{f1faee}

\newcommand{\change}[1]{\textcolor{black}{#1}}

\newcommand\copyrighttext{%
  \footnotesize \textcopyright 2022 IEEE. Personal use of this material is permitted. Permission from IEEE must be obtained for all other uses, in any current or future media, including reprinting/republishing this material for advertising or promotional purposes, creating new collective works, for resale or redistribution to servers or lists, or reuse of any copyrighted component of this work in other works. DOI:10.1109/LRA.2022.3150505.}
\newcommand\copyrightnotice{%
\begin{tikzpicture}[remember picture,overlay]
\node[anchor=south] at (current page.south) {\fbox{\parbox{\dimexpr\textwidth-\fboxsep-\fboxrule\relax}{\copyrighttext}}};
\end{tikzpicture}%
}

\title{EchoVPR: Echo State Networks for \\Visual Place Recognition
}

\author{An{\i}l {\"O}zdemir$^{1,\dag}$, Mark Scerri$^{1,\dag}$,  Andrew B. Barron$^{2}$, Andrew Philippides$^{3}$, \\Michael Mangan$^{1,*}$, Eleni Vasilaki$^{1,4,*}$, and Luca Manneschi$^{1,*}$%
\thanks{Manuscript received: September 9, 2021; Revised December 3, 2021; Accepted January 20, 2022.}
\thanks{This paper was recommended for publication by Editor Eric Marchand upon evaluation of the Associate Editor and Reviewers' comments.}%
\thanks{This work was supported by Engineering and Physical Sciences Research Council under Grant EP/P006094/1, EP/S030964/1, EP/S009647/1 and EP/V006339/1, and by Templeton World Charity Foundation Grant \textnumero~0539.} 
\thanks{$^{1}$An{\i}l {\"O}zdemir, Mark Scerri, Michael Mangan, Eleni Vasilaki, and Luca Manneschi are with Department of Computer Science, The University of Sheffield, UK
        {\tt\footnotesize e.vasilaki@sheffield.ac.uk}}%
\thanks{$^{2} $Andrew B. Barron is with Department of Biological Sciences, Macquarie University, Australia}%
\thanks{$^{3} $Andrew Philippides is with School of Engineering and Informatics, University of Sussex, UK}%
\thanks{$^{4} $Eleni Vasilaki is Institute for Neuroinformatics, University of Zurich and ETH Zurich, Switzerland}%
\thanks{$^{\dag}$These authors have equally contributed to the work.}%
\thanks{$^{*}$Joint senior authors.}%
}

\begin{document}

\maketitle
\copyrightnotice

\begin{abstract}
Recognising previously visited locations is an important, but unsolved, task in autonomous navigation. 
Current visual place recognition (VPR) benchmarks typically challenge models to recover the position of a query image (or images) from sequential datasets that include both spatial and temporal components.  
Recently, Echo State Network (ESN) varieties have proven particularly powerful at solving machine learning tasks that require spatio-temporal modelling.
These networks are simple, yet powerful neural architectures that---exhibiting memory over multiple time-scales and non-linear high-dimensional representations---can discover temporal relations in the data while still maintaining linearity in the learning time.
In this paper, we present a series of ESNs and analyse their applicability to the VPR problem.
We report that the addition of ESNs to pre-processed convolutional neural networks led to a dramatic boost in performance in comparison to non-recurrent networks in \change{five out of six} standard benchmarks (GardensPoint, SPEDTest, ESSEX3IN1, \change{Oxford RobotCar, and Nordland)}, demonstrating that ESNs are able to capture the temporal structure inherent in VPR problems. 
Moreover, we show that \change{models that include} ESNs can outperform class-leading VPR models which also exploit the sequential dynamics of the data. 
Finally, our results demonstrate that ESNs improve generalisation abilities, robustness, and accuracy further supporting their suitability to VPR applications.

\end{abstract}


\section{Introduction}
\label{sec:introduction}

\IEEEPARstart{V}{isual} Place Recognition (VPR) challenges algorithms to recognise previously visited places despite changes in appearance caused by illuminance, viewpoint, and weather conditions \change{(for reviews see~\cite{lowry2015visual, masone2021survey, zhang2021visual})}. 
Unlike in many machine learning tasks, typical VPR benchmark\change{s challenge models to learn image locations following a single route traversal, which are then} compared with data during another route traversal. 
\change{Thus,} there are very few examples to learn from (typically only the images within a few metres of the correct location) making the task \change{particularly} challenging.

\change{One approach is to recognise places based on matching single views using image processing methods to remove the variance between datasets. For instance, models have been developed that use different image descriptors to obtain meaningful image representations that are robust to visual change (e.g. DenseVLAD~\cite{torii2015place}, NetVLAD~\cite{arandjelovic2016netvlad}, AMOSNet~\cite{chen2017deep}, SuperGlue \cite{sarlin2020superglue}, DELG \cite{cao2020unifying}, Patch-NetVLAD \cite{hausler2021patch}, and HEAPUtil \cite{keetha2021hierarchical}).}
While matching single images is successful in many scenarios, it can suffer from the effects of aliasing, individual image corruption, or sampling mismatches between datasets (e.g. it is challenging to ensure that images sampled along the same route precisely overlap).

\change{Milford and Wyeth~\cite{milford2012seqslam} were the first to demonstrate that such issues could be overcome by matching sequences of images using a global search to overcome individual image mismatches.}
\change{This has led to a family of models that improve VPR performance by exploiting the temporal relationships inherent in images sampled along routes~\cite{milford2012seqslam, milford2013vision, hansen2014visual, kagioulis2020insect, zhu2020spatio, chancan2020hybrid, garg2021seqnet}.}
While \change{achieving state-of-the-art performance on challenging real-world datasets (e.g. 
Oxford RobotCar~\cite{maddern2017year}, Extended CMU Seasons~\cite{sattler2018benchmarking}, Pittsburgh~\cite{torii2013visual}, Tokyo24/7~\cite{torii2015place}, Nordland~\cite{sunderhauf2013we}, Mapillary Street Level Sequences~\cite{warburg2020mapillary}),  such models} often \change{include an explicit encoding of non-visual information to limit the image search space, require a cache of input images for global searches, and can be computationally expensive in both learning and deployment phases. Neither of these features are desirable for autonomous robots that may have limited computational resources} and external sensing capabilities.  

\begin{figure*}[t]
    \centering
    \includegraphics[width=0.95\textwidth]{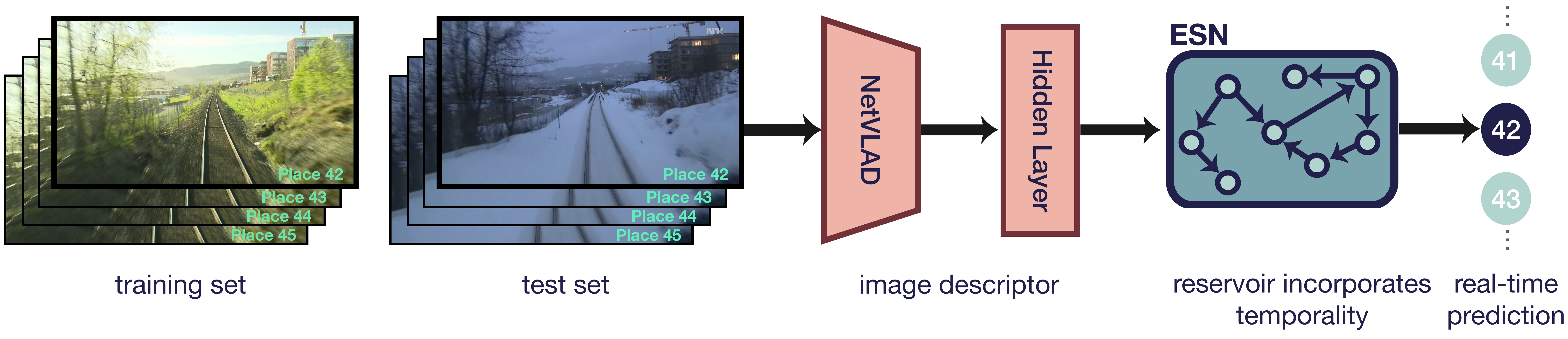} 
    \caption{
    \textbf{An illustration of EchoVPR framework.} 
    Echo State Networks (ESN) incorporate temporality while still maintaining real-time prediction capability, which is a key feature for a robotic system in real-world applications.
    Given an input image at a time (from snowy Nordland~\cite{sunderhauf2013we} in this example), an image descriptor (class-leading NetVLAD~\cite{arandjelovic2016netvlad} \change{via a hidden layer, see Section~\ref{sec:methods:benchmarks-preprocessing})} provides a meaningful representation to the ESN to update the \textit{fixed} reservoir. 
    }
    \label{fig:title-figure}
\end{figure*}

Echo State Networks (ESN)~\cite{jaeger2007optimization} are a class of \change{computationally efficient} recurrent neural networks, ideally suited to addressing VPR problems without the need for additional support cues or input data caching (see Fig.~\ref{fig:title-figure}).
ESNs are a subset of reservoir computing models in which the reservoir neurons possess fixed, random and recurrent interconnections that sustain recent memories, i.e. \textit{echoes}~\cite{jaeger2001echo}, with the practical benefit that only the output layer weights require training.
ESNs thus act as a temporal kernel~\cite{hermans2012recurrent} over a variety of time-scales, creating a form of working memory dispensing of the need for input caching. 
\change{ESNs} have excelled when applied to problems that involve sequential data including dynamical system predictions~\cite{li2012chaotic, deihimi2012application}, robotic motion and navigation tasks~\cite{ploger2003echo, ishu2004identification, hartland2007using}.
\change{Hence, they appear well-suited to the VPR problem but this hypothesis is yet to be tested.} Recently ESNs have been augmented by novel techniques e.g., ~\cite{manneschi2021SPARCE,manneschi2021exploiting}. \change{Among these, in the present work we focus on SPARCE, which stands for Sparse Reservoir Computing.}

\change{The contributions of this work are:
\begin{itemize}
    \item a novel approach that combines state-of-the-art image descriptors augmented with ESNs to exploit temporal dynamics of image sequences;  
    \item validation of the proposed models in four evaluation datasets that present different input variances (e.g. indoor/outdoor, day/night, and viewpoint);
    \item benchmarking against contemporary single- and sequential-view based models in challenging, long-range VPR datasets sampled from moving vehicles across illumination and seasonal changes; 
    \item assessment of the models' capacity to recognise routes from a random start-point (akin to kidnapped robot problem).
\end{itemize}
}

\section{Methods}
\label{sec:methods}

\subsection{Problem Formulation}
\label{sec:methods:problem-formulation}

VPR algorithms are provided with a sequence of places (in \change{the} form of images) sampled along a route.
They are \change{then} asked to correctly match (within an acceptable threshold) the places by the image key-frames along the same route at a different time. 
The input data is composed of videos where the network has to correctly infer the location, i.e. the image key-frame that is processed at the considered time. 
In all the tasks there are at least two sequences of images, one used as a training set (i.e. reference) and the other used as a test set (i.e. query), acquired by visiting the same locations and following the same path twice.
Even though there is a one-to-one mapping between training and test samples, the latter is acquired by visiting the locations at different times, leading to differences in visual appearances, such as seasonal or illuminance as well as viewpoint changes.
Perfect matching is not always possible, hence, there can be a tolerance term that allow\change{s} close match\change{es} to be accepted. A match is considered successful, if $||\text{reference} - \text{query}|| \leq \text{tolerance}$.

In our specific implementation, we consider supervised learning with the ESNs as a predictor, hence, forming a classification problem.
The number of read-out nodes is equal to the number of places, and therefore, specific to the given dataset.
The read-out nodes (the final and only learnable layer) output a probability distribution, $\mathcal{P}_{\text{query}}$, for each given query image.
The prediction (i.e. key-frame of the query) is the number of the read-out node, i.e. $\arg \max \mathcal{P}_{\text{query}}$.

\subsection{Standard Echo State Networks}
\label{sec:methods:ESN}

An ESN is a reservoir of recurrently connected nodes, whose temporal dynamics $\mathbf{x}(t)$ evolves following~\cite{jaeger2007optimization}:
\begin{align}
    \mathbf{x}(t+\delta t) & = (1-\alpha)\mathbf{x}(t) + \alpha f\left( \mathbf{h}(t) \right),  \label{eq:esn-def} \\
    \mathbf{h}(t) & = \gamma \mathbf{W}_{\rm in} \mathbf{s}(t) + \rho \mathbf{W} \mathbf{x}(t), \label{eq:esn-hidden} 
\end{align}
where $\alpha$ is the leakage term and defines the rate of integration of information, $f$ is a non-linear activation function (usually $\operatorname{tanh}$), $\mathbf{s}(t)$ is the input signal, $\mathbf{W}_{\rm in}$ is the input connectivity matrix, which is commonly drawn from a random Gaussian distribution, and $\gamma$ is a multiplicative factor of the external signal. 
The recurrent connectivity $\mathbf{W}$ is a sparse, random and fixed matrix whose eigenvalues are constrained inside the unit circle of the imaginary plane, with a hyper-parameter $\rho$ (usually in the range of $[0,1]$) set to further control the spectral radius. 
\change{Learning occurs via minimisation of a cost function. Only the read-out weights $\mathbf{W}_{\rm out}$ that connect the reservoir neurons $\mathbf{x}$ to the output change.} 
Optimisation of $\mathbf{W}_{\rm out}$  can be accomplished through different techniques \change{such} as ridge regression or iterative gradient descent methods \cite{lukovsevivcius2012practical}. \change{In our work, we use exclusively gradient decent methods.}       

\subsection{\change{Sparse Reservoir Computing}}
\label{sec:method:hier-ESN}

The definition of sparse representations through the SPARCE model~\cite{manneschi2021SPARCE} can enhance the capacity of the reservoir to learn associations by introducing specialised neurons through the definition of learnable thresholds. 
Considering the representation $\mathbf{x}$ from which the read-out is defined, as in Eq.~\eqref{eq:esn-def}, SPARCE consists of the following normalisation operation:
\begin{align}
    x^{\prime}_i &= \operatorname{sign} \big( x_i \big) \operatorname{ReLU} \Big( |x_i| - \theta_i \Big) \label{eq:SPARCE-x} \\
    \theta_i &= P_{n} \big( |\mathbf{x}_i| \big) + \Bar{\theta}_i \label{eq:SPARCE-theta} 
\end{align}
where $i$ is the $i$-th dimension, $\operatorname{sign}$ is the sign function and $\operatorname{ReLU}$ is the rectified linear unit. Of course, the new read-out is defined from $x^{\prime}_i$, that is after the transformation given in Eq.~\eqref{eq:SPARCE-x} and~\eqref{eq:SPARCE-theta}, which leaves the dynamics of the system unaltered and can be easily applied to any reservoir representation. 
The threshold $\theta_i$ is composed of two factors: $P_{n}\big(|\mathbf{x}_i|\big)$, i.e. the $n$-th percentile of $\mathbf{x}_i$, which stands for the distribution of activities of dimension $i$ after the presentation of a number of samples with sufficient statistics, and a learnable part $\Bar{\theta}_i$, which is adapted through gradient descent and is initialised to arbitrarily small values at the beginning of training. 
The percentile $n$ can be considered as an additional \textit{interpretable} hyper-parameter that controls the sparsity level of the network at the start of the training phase\footnote{For different methodologies to estimate the percentile operation, see~\cite{manneschi2021SPARCE}}.   

\subsection{Image Pre-processing}
\label{sec:methods:benchmarks-preprocessing}

Convolutional neural networks (CNN) are the best performing architectures for processing images and discover\change{ing} high-level features from visual data. 
However, they are static and lack temporal dynamics. 
Thus, after a pre-processing module composed of NetVLAD~\cite{arandjelovic2016netvlad}, a pre-trained CNN, we adopted a system composed by ESNs. 
Considering that the reservoir computing paradigm is more effective when the reservoir expands the dimensionality of its corresponding input, we first decreased the dimensionality of NetVLAD output (original dimension is $4096$) by training a feedforward network composed of one hidden layer (with $500$ nodes) on the considered classification task. 
The hidden layer  representation is then considered as the input to the reservoir computing system. With NV we denote the full model consisting of NetVLAD, a hidden layer and an output layer. When NV is combined with another system, e.g., NV-ESN, it means that we have used the trained hidden layer as an input representation to the ESN.  

\subsection{Proposed Architecture}
\label{sec:parch}

The reservoir is then trained to distinguish the different locations, which are processed successively in the natural order of acquisition by the overall architecture. 
The \change{two reservoir computing models} we study are summarised below\footnote{The supplementary material and source-code for the ESN implementations can be found in \url{https://anilozdemir.github.io/EchoVPR/}}:
\begin{itemize}
    \item \textbf{Echo State Network (\texttt{ESN)}}, where learning happens on the output weights only.
    The critical hyper-parameters of the system for the cases studied are $\alpha, \gamma, \eta$ (leakage term, input factor, learning rate). \change{The overall architecture that exploits this reservoir will be named \texttt{NV-ESN} in the rest of the work.}
    \item \textbf{Echo State Network with \change{SPARCE} (\texttt{\change{SPARCE-ESN}})}, where thresholds are applied to the reservoir following Eq.~\eqref{eq:SPARCE-x} and learning occurs on $\Bar{\theta}$ and $\mathbf{W}_{\rm out}$. 
    The hyper-parameters are the same as the standard \texttt{ESN} with the addition of the starting percentile $P_n$ of Eq.~\eqref{eq:SPARCE-theta}. \change{For this case, the overall architecture will be named \texttt{NV-SPARCE-ESN}.}
\end{itemize}   

\subsection{\change{Error Functions}}
\change{Learning of $\mathbf{W}_{\rm out}$ and $\Bar{\theta}$ is accomplished through mini-batches and by minimisation of softmax cross-entropy loss via gradient descent:} 
\begin{equation}
    E = \sum_j^{\rm N_{batch}}\sum_i {y^{\rm target}_{ij}} 
    \operatorname{log} 
    \Bigg( 
    \dfrac{\exp\big(y_{ij}\big)}
    {\sum_i \exp(y_{ij})}
    \Bigg),
    \label{eq:softmax_loss}
\end{equation}
where $\rm N_{batch}$ is the minibatch size, $y$ the output of the neural network, $y^{\rm target}$ the target output, and the \change{indices} $i$ and $j$ correspond to the sample number and to the output node considered.
The models are trained for up to $60$ epochs, i.e. each training image is \change{seen} $60$ times.

\begin{figure*}[t]
    \centering
    \includegraphics[width=0.87\textwidth]{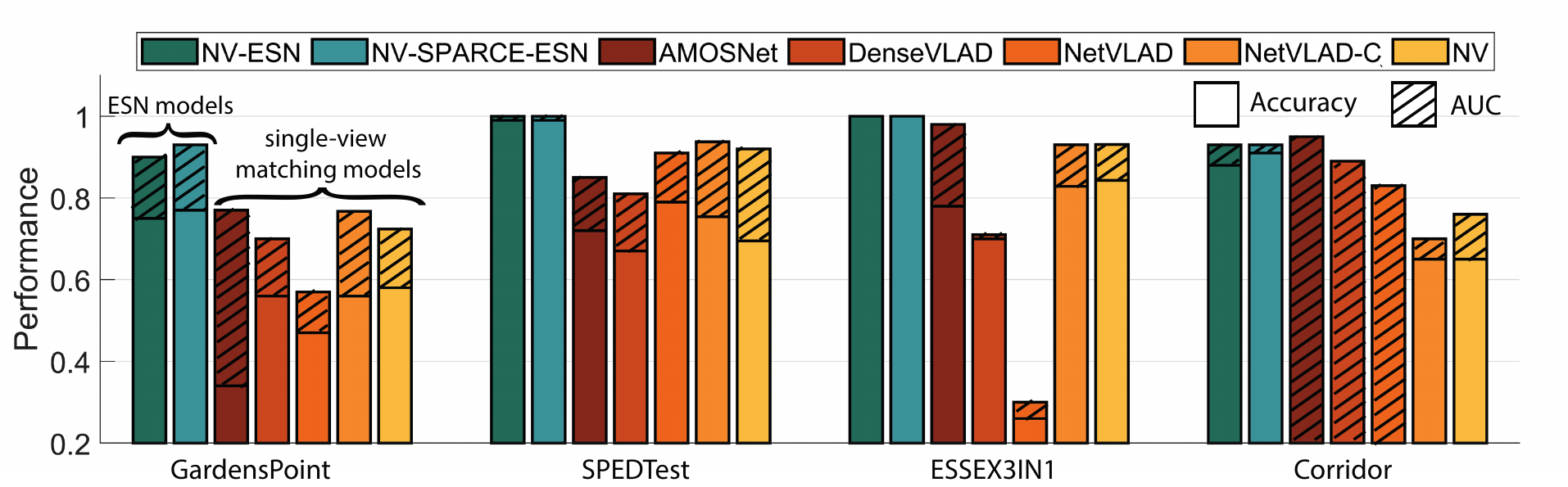} 
    \caption{\textbf{ESN performance compared with single view matching algoirthms in evaluation datasets.} 
    The utilisation of reservoir computing models permits the temporal dynamics of the problem to be captured and improves the performance of CNNs. 
    \change{\texttt{NV-ESN}} and \change{\texttt{NV-SPARCE-ESN}} are shown in blue-green colours, while the performance of static neural networks is reported in red-yellow colours. 
    The performance of AMOSNet, DenseVLAD and NetVLAD were taken from~\cite{zaffar2021vpr}, where image matching was achieved by computing distances among the representation. 
    \change{NetVLAD-C} and \change{NV} correspond to models in which a simple read-out or a hidden layer \change{with a read-out} were trained from the representation of the convolutional network respectively. 
    This was achieved through the minimisation of Eq.~\eqref{eq:softmax_loss} on the specific task considered, similar to the approach used for ESNs.
    The bar plots for our method shows average performance over $20$ trials.
    }
    \label{fig:comparison}
\end{figure*}

\change{For larger datasets}, we used the sigmoid cross-entropy loss as the error function, which led to better performance:
\begin{equation}
    \small
    E=\sum_j^{\rm N_{batch}}\sum_i y^{target}_{ij} 
    \operatorname{log} 
    \change{
    \big[ \sigma(y_{ij}) \big]+(1-y^{target}_{ij}) 
    \operatorname{log} 
    \big[ 
    1-\sigma(y_{ij})
    \big]},
    \label{eq:sigmoid_loss}
\end{equation}
where \change{$\sigma$ is the sigmoid function and the  remaining} terms \change{correspond} to the ones of Eq.~\eqref{eq:softmax_loss}.
A description of the datasets used is provided in Section~\ref{sec:exp:dataset-metrics}.

\section{Experiments}
\label{sec:exp}

\subsection{\change{Datasets And Metrics}}
\label{sec:exp:dataset-metrics}

\change{To allow benchmarking of different models across datasets, the VPR community is converging upon a standardised methodology~\cite{zaffar2021vpr, garg2021your, toft2020long}.
We therefore evaluate model performance using the recommended metrics: prediction \textit{accuracy}, \textit{recall@n} and precision-recall \textit{area-under-curve} (AUC).
}

\change{
Initial analysis was performed on four datasets that challenge models in a variety of conditions: 
\begin{itemize}
    \item \textbf{GardensPoint}~\cite{glover2014day} consists of $200$ indoor, outdoor and natural environments with both viewpoint and conditional changes throughout the dataset. A tolerance of $2$ is accepted. 
    \item \textbf{ESSEX3IN1}~\cite{zaffar2020memorable} consists of $210$ images taken at the university campus and surroundings, focusing on perceptual aliasing and confusing places. There is no tolerance for this dataset.
    \item \textbf{SPEDTest}~\cite{chen2018learning} consists of $607$ low-quality but high-depth images collected from CCTV cameras around the World; it includes environmental changes including variations in weather, seasonal, and illumination conditions. There is no tolerance for this dataset.
    \item \textbf{Corridor}~\cite{milford2013vision} consists of 111 images of an indoor environment - three traverses of the same corridor.
\end{itemize}
} 

\change{Models were then evaluated in two larger datasets captured from moving vehicles that are more representative of the data that would be presented to an autonomous robot:}
\change{
\begin{itemize}
    \item \textbf{Nordland}~\cite{hausler2021patch} comprises of images taken at train traversals in four different seasons in Norway; the viewpoint angle is fixed although there is a high weather, seasonal and illumination variability. Each traversal consists of $27592$ images. A tolerance of $10$ is acceptable---the same as the models~\cite{chancan2020hybrid,hausler2021patch,garg2021seqnet} we compare against (see Section~\ref{sec:results:nordland-sequential-comparison} for more details).
    \item \textbf{Oxford RobotCar}~\cite{maddern20171} comprises of images taken from a car travelling a fixed 10 km route around Oxford, UK at different times of the day across seasons. Each traversal consists of roughly $30000$ images.
\end{itemize}
}

\change{To allow direct comparison of ESNs with different published results, we also consider a subset of the \textbf{Nordland} dataset composed of $1000$ images as in \cite{chancan2020hybrid}, and then the full dataset as in \cite{hausler2021patch}. 
For the \textbf{Oxford RobotCar} dataset \cite{maddern20171}, we used 4599 and 4550 images captured during the day (2015-03-17-11-08-44) and at night (2014-12-16-18-44-24) sampled at $2$m apart for training and testing respectively, with a tolerance of $20$m, in a similar manner to~\cite{garg2021seqnet}.
} 

\subsection{\change{Benchmarking Process}}
\label{sec:exp:benchmarking}
\change{We compared \texttt{NV-ESN} and \texttt{NV-SPARCE-ESN} (see Section~\ref{sec:parch}) on the smaller datasets (GardensPoint, SPEDTest, ESSEX3IN1) to the following models, taken from the literature: AMOSNet \cite{chen2017deep}, DenseVLAD \cite{torii2015place} and NetVLAD \cite{arandjelovic2016netvlad}. 
We also used our own NetVLAD variants, NetVLAD-C, which consists of NetVLAD with an output layer trained as a classifier, and NV (see Section~\ref{sec:methods:benchmarks-preprocessing}). }

\change{For the Nordland dataset, we compare our methods with NetVLAD~\cite{arandjelovic2016netvlad}, SuperGlue~\cite{sarlin2020superglue}, DELG global~\cite{cao2020unifying}, DELG local~\cite{cao2020unifying} and Patch-NetVLAD~\cite{hausler2021patch}. 
The latter uses NetVLAD in combination with local features (Patch). 
We were able to combine our ESNs variants with local features similar to Patch-NetVLAD. 
In particular, after classification of the top $100$ locations performed by \texttt{NV-ESN} or \texttt{NV-SPARCE-ESN}, we used such local features to select the final classified location. 
For simplicity, we named the resulting models \texttt{PatchL-ESN} and \texttt{PatchL-SPARCE-ESN} for the case with or without SPARCE respectively, where \texttt{L} stands for ``Light''. 
The ``Light'' model refers to the variant Single-Spatial-Patch-NetVLAD as defined in~\cite{hausler2021patch} which employs a simple spatial verification method applied on a single patch of size $5$ and $512$ PCA dimensions. 
In contrast the results that are reported from Patch-NetVLAD~\cite{hausler2021patch} utilises the variant Multi-RANSAC-Patch-NetVLAD which employs a more complex spatial scoring method applied on a fusion of multiple patches of sizes ($2$, $5$ and $8$) and $4096$ PCA dimensions. This selection process among the top $100$ images is analogous to the one in the original paper~\cite{hausler2021patch}.
}

\change{Again on Nordland, we compared \texttt{NV-ESN} to sequential models. 
More specifically, we considered three NetVLAD variants (NetVLAD+Smoothing, NetVLAD+Delta and Netvlad+SeqMatch), SeqNet($S_{5}$) and HVPR($S_5$ to $S_1$). 
For the description of all these models see~\cite{garg2021seqnet}. 
SeqNet, in particular, is a recently formulated sequential model where authors used the images of the Nordland dataset captured during summer and winter as training set, and then tested the model on previously unseen locations during Spring and Fall. 
In~\cite{garg2021seqnet}, the generalisation on unseen locations is possible because of the computation of distances among images and of the minimisation of a triplet loss function. 
Considering that we are training a classifier, where each image corresponds to a different output node similarly to~\cite{chancan2020hybrid}, it is difficult to make a direct comparison. 
Instead, we formulated a methodology inspired by the one adopted for SeqNet~\cite{garg2021seqnet}. 
We selected single images or pairs of consecutive images randomly from the summer and winter datasets and excluded them from the training. 
We used these images for validation and images captured at the same locations, but during Spring and Fall, for testing. 
In this way, our models are tested on images for which training was absent. }

\change{For the reduced version of the Nordland dataset, we compared our methods \texttt{NV-ESN} and \texttt{NV-SPARCE-ESN} to FlyNet+CANN and FlyNet+RNN. 
In that case the comparison was straightforward as all networks are classifiers presuming each place is a class.}

\change{We also compared \texttt{NV-ESN}, \texttt{PatchL-ESN}, \texttt{NV-SPARCE-ESN} and \texttt{PatchL-SPARCE-ESN} on the Oxford RobotCar dataset with NetVLAD and Patch-NetVLAD, which were not available from the literature. 
For these two models, we produced results using the source-code provided by the authors.}

\subsection{Hyper-parameter Tuning On ESNs}
\label{sec:exp:training}

The lack of a validation set for the considered tasks makes hyper-parameter selection challenging. 
This difficulty is \change{increased} by the small number of samples in the training set (i.e. one sample per place) and by the major statistical differences between \change{the} training and test data. 
In particular, the seasonal \change{differences} in the acquisition of reference and query data lead to the possible presence or absence of snow and shifts in \change{colour intensity}. 
In our preliminary experiments, different hyper-parameters \change{often reached perfect accuracy (i.e. $100\%$) on the training set and had reduced performance on the test set. For this reason, a validation test was required.}

We believe that there is a lack of clarity in previous research works regarding the definition of a methodology to overcome the problem of hyper-parameter selection. We\change{ therefore} tuned the hyper-parameters of the reservoir by using a small percentage (i.e. $10\%$) of samples of the test set as validation. 
In other words, while the read-out was always optimised from \change{training} samples, hyper-parameters were optimised through grid search over the performance achieved on $10\%$ of the \change{test} data. 
\change{We did not estimate performance on validation data when computing scores on the test dataset.}
Being aware of the limitations of this methodology, we show how it is possible to use the test set of one task as validation for another task with little performance los\change{s}, \change{demonstrating that the model performs well if the hyper-parameters were selected to be robust to non-excessive statistical changes} (see Section~\ref{sec:results:robustness}).

\section{Results}
\label{sec:results}

\subsection{Assessing ESNs On Visual Place Recognition}
\label{sec:results:comparison}

The performance of \change{the \texttt{NV-ESN} and \texttt{NV-SPARCE-ESN}} \change{models} were first evaluated in \change{four} datasets (GardensPoint, SPEDTest, ESSEX3IN1 and \change{Corridor}). Fig.~\ref{fig:comparison} shows that both ESN variants outperform single-view matching models (including NetVLAD with read-out and hidden layers) in all four conditions (with the exception of AMOSNet in Corridor).   
The \change{\texttt{NV-ESN}} achieves mean accuracy scores of $0.75$, $0.99$, $1.0$ \change{and 0.88} and mean AUC scores of $0.9$, $1.0$, $1.0$ \change{and $0.93$}.
The addition of the \change{\texttt{NV-SPARCE}} layer provides additional improvement with accuracy scores of $0.77$, $0.99$, $1.0$, \change{and $0.91$} and mean AUC scores of $0.93$, $1.0$, $1.0$ \change{and $0.93$}.
\change{The ESNs used $N=1000$ reservoir neurons.}

\subsection{\change{Robustness Of Hyper-parameters}}
\label{sec:results:robustness}

\begin{figure}
    \centering
    \includegraphics[width=0.41\textwidth]{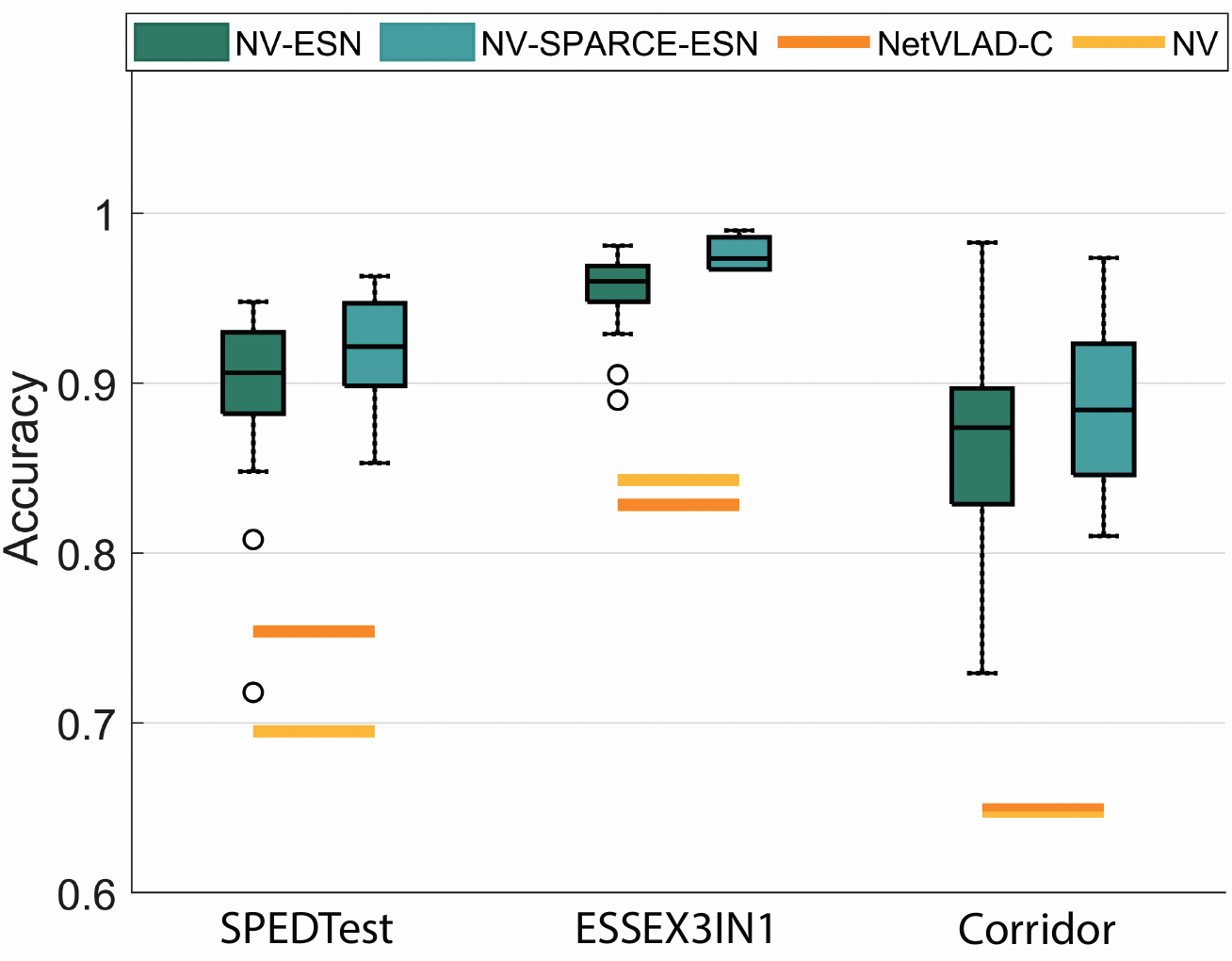} 
    \caption{
    \textbf{\change{Robustness of hyper-parameters.}} 
    \change{Performance is maintained} despite using the hyper-parameters optimised for a different dataset (GardensPoint).
    \change{The two} variants of ESN are well above the accuracy achieved by NetVLAD-C and NV (horizontal lines).
    The box plots represent the distribution of $20$ trials.
    }
    \label{fig:generalisability}
\end{figure}

We also analysed the sensitivity of the ESNs with respect to hyper-parameter selection.
Fig.~\ref{fig:generalisability} shows accuracy scores for hyper-parameters tuned by training the models on GardensPoint and maintaining them when training in SPEDTest, ESSEX3IN1 \change{and Corridor}. 
The reason we chose the hyper-parameters \change{for} GardensPoint is that generalisation is more likely to occur when the baseline task is more complex than the new tasks to which it is applied. 
Indeed, richer and more difficult datasets can lead neural networks to discover high-level features that are transferable to simpler datasets, while the contrary is difficult. 
Fig.~\ref{fig:generalisability} demonstrates how, even with sub-optimal hyper-parameters, the introduction of ESNs leads to higher performance in comparison to single-view matching models: \change{NetVLAD-C and NV}.
Moreover, the performance remains above \change{$86\%$} for both accuracy and AUC compared to the virtually perfect scores achieved when hyper-parameters were tuned using the same dataset (see Fig~\ref{fig:comparison}).

\begin{table}[t]
\centering
\caption{
\change{
\textbf{ESNs outperform state-of-the-art single view matching models in the challenging Nordland and Oxford RobotCar benchmarks.} The larger datasets pose a greater challenge for algorithms with an order of magnitude more images and extreme variance between training and test sets (seasonal and day/night). Best results are found for \texttt{PatchL-ESN} in Nordland, and \texttt{PatchL-SPARCE-ESN} in Oxford RobotCar. }
}
\resizebox{0.49\textwidth}{!}{
\begin{tabular}{|l |c c c| r r r |}
\hline
      & \multicolumn{3}{c|}{Nordland} & \multicolumn{3}{c|}{Oxford RobotCar} \\
      & \multicolumn{3}{c|}{\scriptsize{Summer vs Winter}} & \multicolumn{3}{c|}{\scriptsize{Day vs Night}} \\
 \textbf{Method}       &  R@1 & R@5 & R@10 &  R@1 & R@5 & R@10  \\
  \hline
 NetVLAD \cite{arandjelovic2016netvlad}      & $13.0$ & $20.6$ & $25.0$ & $ 31.2$ & $48.6$ & $58.3$   \\
 SuperGlue \cite{sarlin2020superglue}    & $29.1$ & $33.4$ & $35.0$ & $-$ & $-$ & $-$  \\
 DELG global \cite{cao2020unifying}  & $23.4$ & $35.4$ & $41.7$ & $-$ & $-$ & $-$  \\
 DELG local \cite{cao2020unifying}   & $60.1$ & $63.5$ & $64.6$ & $-$ & $-$ & $-$ \\
 Patch-NetVLAD \cite{hausler2021patch} & $44.9$ & $50.2$ & $52.2$ & $80.7$ & $84.8$ & $86.4$  \\
\hline
\texttt{\textbf{NV-ESN}}   & $47.5$ & $61.0$ & $66.4$ & $44.0$ & $58.1$ & $65.5$  \\
\texttt{\textbf{NV-SPARCE-ESN}}   & $46.7$ & $60.3 $ & $65.9 $ & $80.4 $ & $87.9$ & $91.5$  \\
\texttt{\textbf{PatchL-ESN}}  & $\mathbf{66.4} $ & $\mathbf{76.0} $ & $\mathbf{78.7} $ & $75.9$ & $85.2 $ & $87.7 $ \\
\texttt{\textbf{PatchL-SPARCE-ESN}}  & $66.3$ & $75.7$ & $78.6$ & $\mathbf{88.8}$ & $\mathbf{97.0} $ & $\mathbf{98.2} $ \\
\hline

\end{tabular}}
\label{tab1}
\end{table}

\subsection{\change{Performance On Larger Datasets}}
\label{sec:results:more-complex}

\change{To assess the scalability of the ESNs, and to compare their performance to state-of-the-art models, performance was analysed in the full Nordland and Oxford RobotCar datasets as in \cite{garg2021seqnet}. 
The methodology adopted is the same as in Section~\ref{sec:results}, but the number of reservoir nodes in the ESN is increased to $8000$ and $6000$ for the Nordland and the Oxford RobotCar datasets, respectively. 
Results presented in Table~\ref{tab1}, show that best performance was achieved for PatchL-ESN in the Nordland dataset, when compared with state-of-the-art single view matching models.
}
\change{
We would like to note that, in our models we have used exclusively the ``light'' version of Patch-NetVLAD for computational efficiency, while the performance reported in Table~\ref{tab1} of the original Patch-NetVLAD exploited its most complex version. 
Considering the results for Nordland (summer vs winter) and Oxford RobotCar (day vs night), it is evident that the temporal features captured by the ESNs cause it to reach competitive performance. 
\texttt{NV-ESN} reports recall@n that are higher than the original Patch-NetVLAD for the Nordland dataset, while \texttt{NV-SPARCE-ESN} is surprisingly successful on the Oxford RobotCar dataset, with performance that is inferior only to its more complex variant \texttt{PatchL-SPARCE-ESN} and comparable to Patch-NetVLAD. 
The utilisation of the local features from~\cite{hausler2021patch} further improve these results, and the model \texttt{PatchL+SPARCE-ESN} is the best performing network overall. 
Indeed, \texttt{PatchL+SPARCE-ESN} reports the highest score for Oxford RobotCar and it is only slightly worse than \texttt{PatchL-ESN} for Nordland. 
It is likely that these results might improve by using the complete variant of Patch-NetVLAD. 
Here, we chose to keep the model as simple as possible, given that it still outperforms the other techniques.  
Thus, ESNs constitute an addition to other neural architectures that can lead to considerable performance improvements and to state-of-the-art results.
}

\subsection{\change{Robustness With Respect To Start-Point}}
\label{sec:results:robustness-to-start-point}

\begin{figure}[t]
 \centering
    \includegraphics[width=0.42\textwidth]{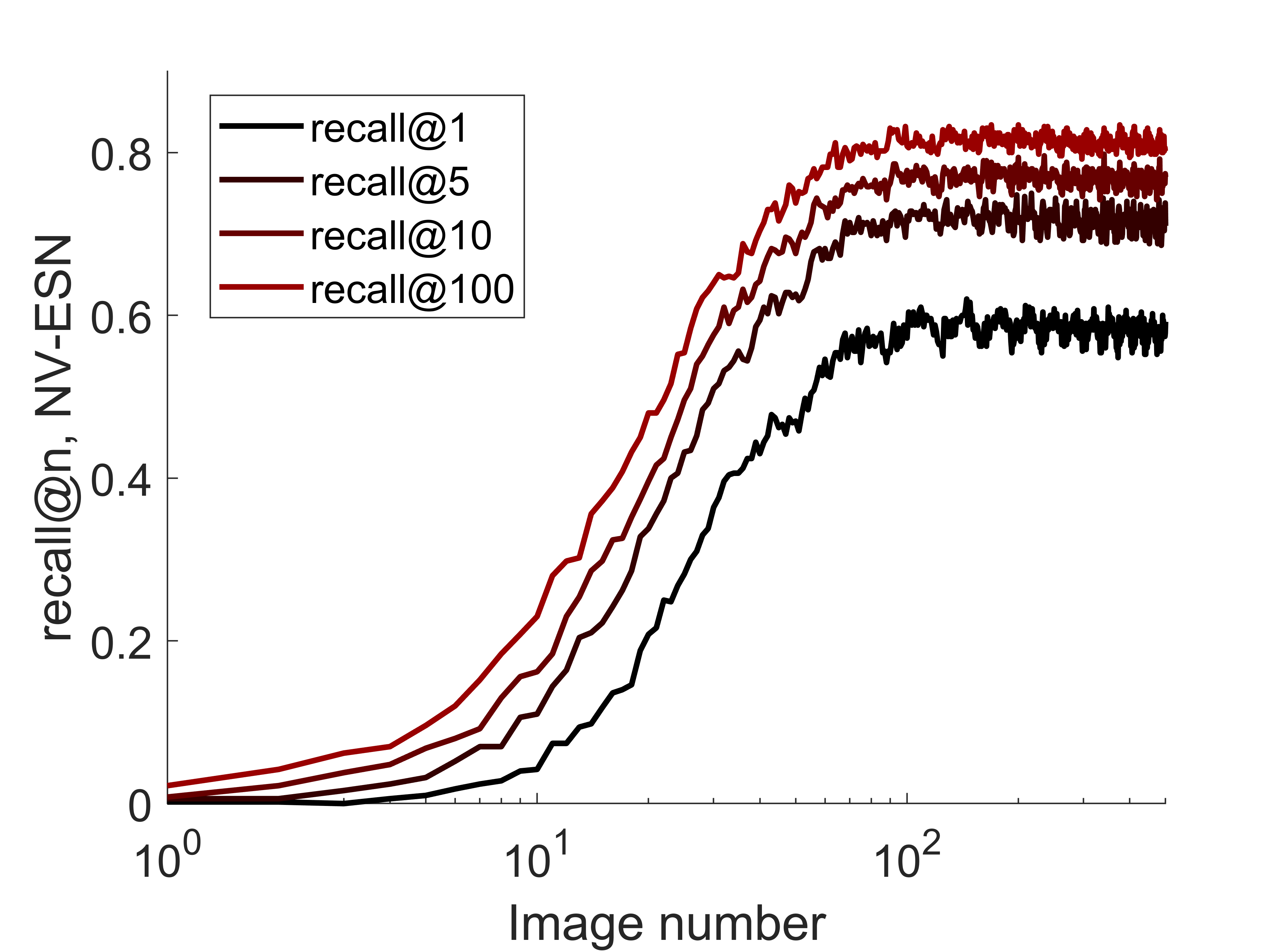}
    \caption{
    \change{
        \textbf{Robustness with respect to start-point.}
        Performance of \texttt{NV-ESN} as the image number from a starting location increases. 
        The trends are averaged across $500$ different starting locations of the Winter Nordland datasets.}}
  \label{fig:starting_locations}
\end{figure}

\change{We investigate if the proposed models satisfy two conditions that are necessary for a potential implementation on a robot. First, the model should maintain high performance if we start from different locations of the input sequence. Second, the amount of temporal information that needs to be processed before the system can reach satisfactory results should be low. Fig.~\ref{fig:starting_locations} shows the recall@n of an ESN as the number of images being processed increases from the starting location. The recall@n are averaged across different possible initial locations and computed on the test set of the Nordland dataset. The average performance increases quickly and steadily, showing that the model needs to process in the order of $50$ images before the trends converge. The fact that the performance of Fig.~\ref{fig:starting_locations} is obtained by averaging from various starting points further demonstrates the robustness of the model with respect to variation of the initial location.
}

\subsection{Comparing ESN With Sequential VPR Models}
\label{sec:results:nordland-sequential-comparison}

\begin{figure}[t]
 \centering
    \includegraphics[width=0.4\textwidth]{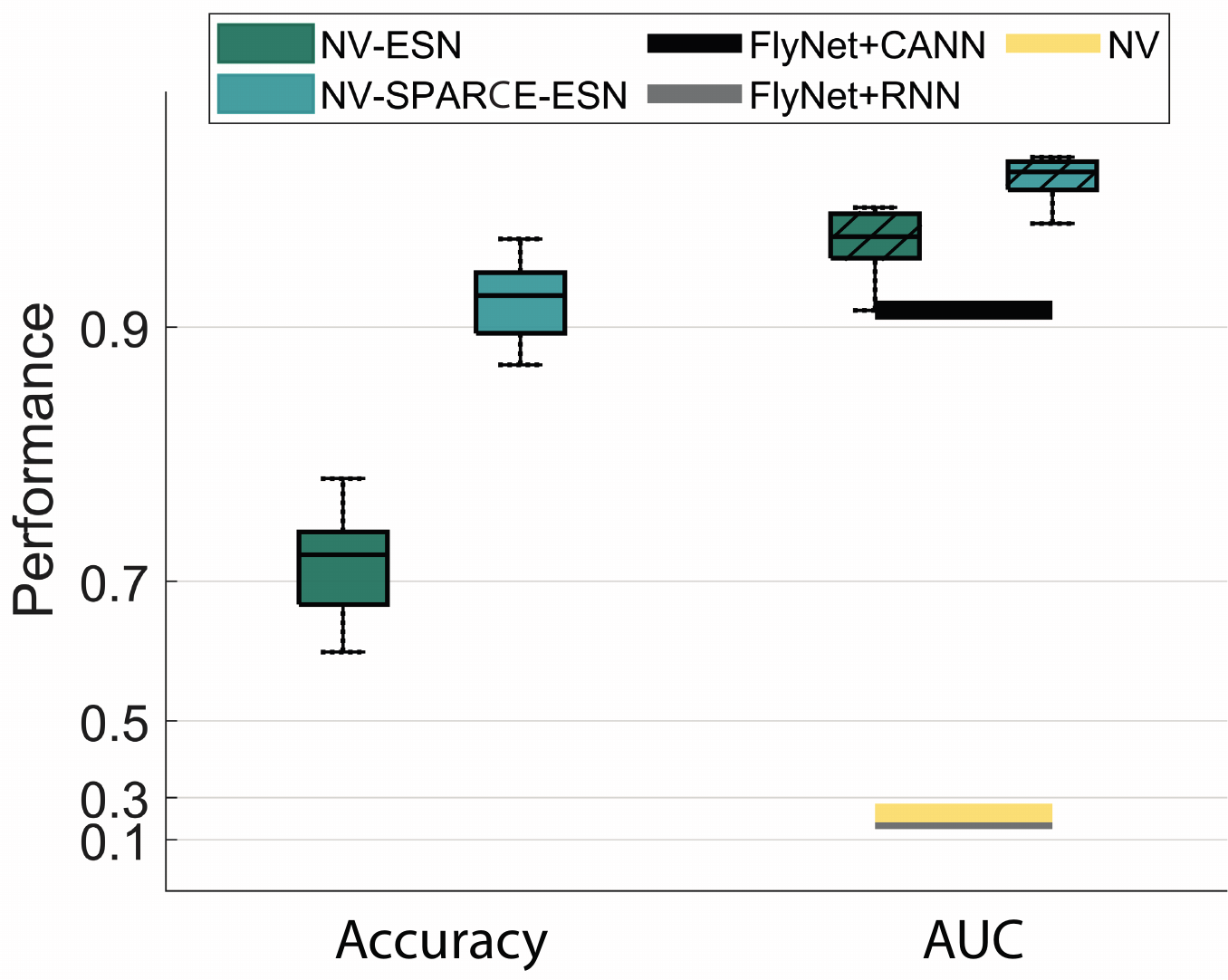}
    \caption{
        \textbf{Direct comparison to sequential classification models in the Nordland dataset.}
        The \change{\texttt{NV-ESN}} model and in particular \change{\texttt{NV-SPARCE-ESN}}, outperform FlyNet+RNN and FlyNet+CANN, taken from~\cite{chancan2020hybrid} and NV when compared with reported AUC scores.
        The horizontal lines report the performance of FlyNet and NV models. The box plots represent the distribution of $20$ trials.}
   
  \label{fig:nordland}
\end{figure}

In this section, we benchmark the performance of ESNs against 
sequence matching VPR models. 
\change{ The majority of sequential models for VPR benchmarks adopt different learning and testing methodologies, and it is consequently difficult to make direct comparisons to previously published results. 
However, it is possible to directly compare our networks with  with two} models recently reported to achieve great performance~\cite{chancan2020hybrid} in \change{a subset of the} Nordland dataset~\cite{sunderhauf2013we} \change{composed by 1000 images}.  
Both \change{previous} models use a bio-inspired feedforward neural network (FlyNet) to encode visual information and either a recurrent neural network (RNN) or a continuous attractor \change{neural} network (CANN) to introduce temporality.
Fig.~\ref{fig:nordland} shows accuracy scores of $0.72$ and $0.92$ for the standard \change{\texttt{NV-ESN}} and \change{\texttt{NV-SPARCE-ESN}} respectively.
For the AUC test, \change{\texttt{NV-ESN}} achieves scores of $0.95$, with \change{\texttt{NV-SPARCE-ESN}} improving results to $0.98$.
This compares favourably to both static view matching models (e.g. NetVLAD+HL) which score $0.24$, and sequential models which score $0.21$ (FlyNet+RNN) and $0.91$ (FlyNet+CANN). 
\change{While it is not possible to directly and fairly compare with models such as SeqNet~\cite{garg2021seqnet} that use distances between images, we have devised a methodology that allows us to provide scores of our methods in classifying previously unseen data.
}

\change{
The performance obtained for the larger version of the Nordland dataset are shown in Table~\ref{tab2} (\textit{summer-winter vs spring-fall}). 
The two results reported for each recall@n correspond to the case where single, or consecutive image pairs, were randomly selected for testing respectively (see Section~\ref{sec:exp:benchmarking} for more details). 
In the latter case, because the connections to adjacent pairs of output neurons are not trained, the performance degrades.
However, even in a learning and testing paradigm not completely suited to our model, the proposed networks are able to generalise. 
This additional comparison between ESNs and previously published results (Table~\ref{tab2}) confirms that the former can achieve high performance in relation to sequential techniques. 
Moreover, the memory of ESNs is solely reliant on the internal dynamics, while the vast majority of other sequential models rely on a direct comparison among sequences, which need to be stored in memory.} 

\begin{table}[t]
\centering
\caption{
\change{
\textbf{Indicative comparison of ESNs to published results with sequential models in the Nordland dataset.} We note that the NV-ESN cannot be evaluated in the same way as the other models so a direct comparison is not appropriate. 
}}
\resizebox{.49\textwidth}{!}{
\begin{tabular}{|l |c c c| }
\hline
      &  \multicolumn{3}{c|}{Nordland} \\
      &  \multicolumn{3}{c|}{\scriptsize{Summer-Winter vs Spring-Fall}}\\

 \textbf{Method}  &  R@1 & R@5 & R@20 \\
  \hline
  NetVLAD+Smoothing \cite{garg2021seqnet}& $44.0$ & $59.0$ & $72.0$  \\
  NetVLAD+Delta \cite{garg2021seqnet} & $56.0$ & $70.0$ & $80.0$ \\
  NetVLAD+SeqMatch \cite{garg2021seqnet}& $61.0$ & $71.0$ & $78.0$ \\
  SeqNet ($S_5$) \cite{garg2021seqnet} & $79.0$ & $90.0$ & $94.0$\\
  HVPR ($S_5$ to $S_1$) \cite{garg2021seqnet} & $79.0$ & $89.0$ & $94.0$ \\
\hline
\hline
NV-ESN & $82.7/77.6$ &$91.2/85.9$  &$95.1/90.9$  \\
\hline

\end{tabular}}
\label{tab2}
\end{table}

\section{Conclusions}
\label{sec:conclusions}

Here we have demonstrated the viability of ESNs as a solution to the VPR problem.
\change{Both ESN variants achieve comparable or better performance than single-view matching models in four evaluation datasets. On larger datasets such as full Nordland and Oxford RobotCar, the two variants, combined with local features, were the best models. 
With respect to sequential models, where a fair comparison was permitted, two ESN variants achieved performance above/equal to the class-leading results on the reduced Nordland dataset. While performance is comparable, we note that FlyNet~\cite{chancan2020hybrid} has many fewer parameters. However, the ESN does not require images to be cached during multiple comparisons or any external cues.}
\change{A comparison with sequential models is, in general, difficult due to their use of distances that allow unseen images to be compared. In this sense, these sequential models have an advantage at the cost that caching images is required in order to recognise a new place.}

\change{The addition of SPARCE to the standard ESN improved performance considerably in almost all cases, showing how the introduction of sparse representations can efficiently help the classification process. 
An exception was the full Nordland dataset, where we did not find any advantage by adding this technique, and the reasons are worthy of further investigation.}

\change{When it comes to ESN advances, an intriguing future course of action is to take inspiration from invertebrate mini-brains. 
An interesting point being that they} possess analogous structural motifs of both deep and shallow ESNs.
A simple example is the insect mushroom body.
This is considered the cognitive centre of the insect brain~\cite{menzel2001cognitive} and is necessary for learning relationships sequences and patterns in honey bees~\cite{menzel2001cognitive, cope2018abstract}.
Structurally the mushroom body is a three-layer network with a compact input layer, an expanded middle layer of inter-neurons called Kenyon cells (KC), and a small layer of output neurons~\cite{fahrbach2006structure}.
The connections between the KC and output neurons are plastic and modified by learning~\cite{gerber2004engram}, and there are chemical and electrical synapses between the KC~\cite{zheng2018complete}.
These features are analogous to the recurrent connections in the reservoir layer of an ESN, and it has been hypothesised~\cite{manneschi2021SPARCE} that these recurrent connections in the KC layer could contribute to the reverberant activity of the mushroom body that supports forms of memory~\cite{cognigni2018right}. Further, the technique SPARCE is inspired by the low activity exhibited by the KC. Given the similar structures of ESN and mushroom body, insights gained from neurobiology could help shape the future ESN investigations and in turn, analysis of the optimal structure for VPR could shed light on the function of different brain areas. \change{For instance, in~\cite{manneschi2021exploiting}, hierarchical structures with fast and slow information processing times are investigated, reminiscent of neural processing observed in insect brains.
\change{See \cite{ozdemir2021echovpr} (supplementary materials)} for a preliminary study of such networks.} 

\change{ESN efficiently exploit information in sequences but in practice it is also desirable that places are recognised from a single input image allowing robotics to truly solve the kidnapped robot problem. 
However, in the cases where such methods fail, traversing portions of a familiar path can help to disambiguate input. 
In this respect, our models require ``viewing'' of approximately $50$ images (for the Nordland set) in order to start correctly identifying their location.
}   

ESNs provide a means to exploit such temporal dynamics using only visual data but more powerful variants require tuning of a large number of parameters which may not be possible when only a small amount of training examples are provided.
Other methods~\cite{milford2012seqslam, chancan2020hybrid} have focused on low-parameter models but often require additional cues such as velocity to focus the image search.
Ensemble methods~\cite{hausler2019multi, fischer2020event} that combine these features are emerging that may provide the best of both worlds.

\addtolength{\textheight}{0cm} 


\bibliographystyle{IEEEtran}
\bibliography{references} 

\begin{thebibliography}{10}
\providecommand{\url}[1]{#1}
\csname url@rmstyle\endcsname
\providecommand{\newblock}{\relax}
\providecommand{\bibinfo}[2]{#2}
\providecommand\BIBentrySTDinterwordspacing{\spaceskip=0pt\relax}
\providecommand\BIBentryALTinterwordstretchfactor{4}
\providecommand\BIBentryALTinterwordspacing{\spaceskip=\fontdimen2\font plus
\BIBentryALTinterwordstretchfactor\fontdimen3\font minus
  \fontdimen4\font\relax}
\providecommand\BIBforeignlanguage[2]{{%
\expandafter\ifx\csname l@#1\endcsname\relax
\typeout{** WARNING: IEEEtran.bst: No hyphenation pattern has been}%
\typeout{** loaded for the language `#1'. Using the pattern for}%
\typeout{** the default language instead.}%
\else
\language=\csname l@#1\endcsname
\fi
#2}}

\bibitem{lowry2015visual}
S.~Lowry, N.~S{\"u}nderhauf, P.~Newman, J.~J. Leonard, D.~Cox, P.~Corke, and
  M.~J. Milford, ``Visual place recognition: A survey,'' \emph{IEEE
  Transactions on Robotics}, vol.~32, no.~1, pp. 1--19, 2015.

\bibitem{masone2021survey}
C.~Masone and B.~Caputo, ``A survey on deep visual place recognition,''
  \emph{IEEE Access}, vol.~9, pp. 19\,516--19\,547, 2021.

\bibitem{zhang2021visual}
X.~Zhang, L.~Wang, and Y.~Su, ``Visual place recognition: A survey from deep
  learning perspective,'' \emph{Pattern Recognition}, vol. 113, p. 107760,
  2021.

\bibitem{torii2015place}
A.~Torii, R.~Arandjelovic, J.~Sivic, M.~Okutomi, and T.~Pajdla, ``24/7 place
  recognition by view synthesis,'' in \emph{Proceedings of the IEEE Conference
  on Computer Vision and Pattern Recognition}, 2015, pp. 1808--1817.

\bibitem{arandjelovic2016netvlad}
R.~Arandjelovic, P.~Gronat, A.~Torii, T.~Pajdla, and J.~Sivic, ``{NetVLAD: CNN}
  architecture for weakly supervised place recognition,'' in \emph{Proceedings
  of the IEEE Conference on Computer Vision and Pattern Recognition}, 2016, pp.
  5297--5307.

\bibitem{chen2017deep}
Z.~Chen, A.~Jacobson, N.~S{\"u}nderhauf, B.~Upcroft, L.~Liu, C.~Shen, I.~Reid,
  and M.~Milford, ``Deep learning features at scale for visual place
  recognition,'' in \emph{IEEE International Conference on Robotics and
  Automation}, 2017, pp. 3223--3230.

\bibitem{sarlin2020superglue}
P.-E. Sarlin, D.~DeTone, T.~Malisiewicz, and A.~Rabinovich, ``Superglue:
  Learning feature matching with graph neural networks,'' in \emph{Proceedings
  of the IEEE/CVF Conference on Computer Vision and Pattern Recognition}, 2020,
  pp. 4938--4947.

\bibitem{cao2020unifying}
B.~Cao, A.~Araujo, and J.~Sim, ``Unifying deep local and global features for
  image search,'' in \emph{European Conference on Computer Vision}.\hskip 1em
  plus 0.5em minus 0.4em\relax Springer, 2020, pp. 726--743.

\bibitem{hausler2021patch}
S.~Hausler, S.~Garg, M.~Xu, M.~Milford, and T.~Fischer, ``Patch-netvlad:
  Multi-scale fusion of locally-global descriptors for place recognition,'' in
  \emph{Proceedings of the IEEE/CVF Conference on Computer Vision and Pattern
  Recognition}, 2021, pp. 14\,141--14\,152.

\bibitem{keetha2021hierarchical}
N.~V. Keetha, M.~Milford, and S.~Garg, ``A hierarchical dual model of
  environment-and place-specific utility for visual place recognition,''
  \emph{IEEE Robotics and Automation Letters}, vol.~6, no.~4, pp. 6969--6976,
  2021.

\bibitem{milford2012seqslam}
M.~J. Milford and G.~F. Wyeth, ``{SeqSLAM: Visual} route-based navigation for
  sunny summer days and stormy winter nights,'' in \emph{IEEE International
  Conference on Robotics and Automation}, 2012, pp. 1643--1649.

\bibitem{milford2013vision}
M.~J. Milford, ``Vision-based place recognition: how low can you go?''
  \emph{The International Journal of Robotics Research}, vol.~32, no.~7, pp.
  766--789, 2013.

\bibitem{hansen2014visual}
P.~Hansen and B.~Browning, ``Visual place recognition using {HMM} sequence
  matching,'' in \emph{2014 IEEE/RSJ International Conference on Intelligent
  Robots and Systems}, 2014, pp. 4549--4555.

\bibitem{kagioulis2020insect}
E.~Kagioulis, A.~Philippides, P.~Graham, J.~C. Knight, and T.~Nowotny, ``Insect
  inspired view based navigation exploiting temporal information,'' in
  \emph{Conference on Biomimetic and Biohybrid Systems}.\hskip 1em plus 0.5em
  minus 0.4em\relax Springer, 2020, pp. 204--216.

\bibitem{zhu2020spatio}
L.~Zhu, M.~Mangan, and B.~Webb, ``Spatio-temporal memory for navigation in a
  mushroom body model,'' in \emph{Conference on Biomimetic and Biohybrid
  Systems}.\hskip 1em plus 0.5em minus 0.4em\relax Springer, 2020, pp.
  415--426.

\bibitem{chancan2020hybrid}
M.~Chanc{\'a}n, L.~Hernandez-Nunez, A.~Narendra, A.~B. Barron, and M.~Milford,
  ``A hybrid compact neural architecture for visual place recognition,''
  \emph{IEEE Robotics and Automation Letters}, vol.~5, no.~2, pp. 993--1000,
  2020.

\bibitem{garg2021seqnet}
S.~Garg and M.~Milford, ``{SeqNet}: Learning descriptors for sequence-based
  hierarchical place recognition,'' \emph{IEEE Robotics and Automation
  Letters}, vol.~6, no.~3, pp. 4305--4312, 2021.

\bibitem{maddern2017year}
W.~Maddern, G.~Pascoe, C.~Linegar, and P.~Newman, ``1 year, 1000 km: The
  {Oxford} {Robotcar} dataset,'' \emph{The International Journal of Robotics
  Research}, vol.~36, no.~1, pp. 3--15, 2017.

\bibitem{sattler2018benchmarking}
T.~Sattler, W.~Maddern, C.~Toft, A.~Torii, L.~Hammarstrand, E.~Stenborg,
  D.~Safari, M.~Okutomi, M.~Pollefeys, J.~Sivic, \emph{et~al.}, ``Benchmarking
  6{DOF} outdoor visual localization in changing conditions,'' in
  \emph{Proceedings of the IEEE Conference on Computer Vision and Pattern
  Recognition}, 2018, pp. 8601--8610.

\bibitem{torii2013visual}
A.~Torii, J.~Sivic, T.~Pajdla, and M.~Okutomi, ``Visual place recognition with
  repetitive structures,'' in \emph{Proceedings of the IEEE conference on
  computer vision and pattern recognition}, 2013, pp. 883--890.

\bibitem{sunderhauf2013we}
N.~S{\"u}nderhauf, P.~Neubert, and P.~Protzel, ``Are we there yet?
  {C}hallenging {SeqSLAM} on a 3000 km journey across all four seasons,'' in
  \emph{Proc. of Workshop on Long-Term Autonomy, IEEE International Conference
  on Robotics and Automation}.\hskip 1em plus 0.5em minus 0.4em\relax Citeseer,
  2013, p. 2013.

\bibitem{warburg2020mapillary}
F.~Warburg, S.~Hauberg, M.~Lopez-Antequera, P.~Gargallo, Y.~Kuang, and
  J.~Civera, ``Mapillary street-level sequences: A dataset for lifelong place
  recognition,'' in \emph{Proceedings of the IEEE/CVF conference on computer
  vision and pattern recognition}, 2020, pp. 2626--2635.

\bibitem{jaeger2007optimization}
H.~Jaeger, M.~Luko{\v{s}}evi{\v{c}}ius, D.~Popovici, and U.~Siewert,
  ``Optimization and applications of {Echo State Networks} with
  leaky-integrator neurons,'' \emph{Neural Networks}, vol.~20, no.~3, pp.
  335--352, 2007.

\bibitem{jaeger2001echo}
H.~Jaeger, ``The “echo state” approach to analysing and training recurrent
  neural networks-with an erratum note,'' \emph{Bonn, Germany: German National
  Research Center for Information Technology GMD Technical Report}, vol. 148,
  no.~34, p.~13, 2001.

\bibitem{hermans2012recurrent}
M.~Hermans and B.~Schrauwen, ``Recurrent kernel machines: Computing with
  infinite echo state networks,'' \emph{Neural Computation}, vol.~24, no.~1,
  pp. 104--133, 2012.

\bibitem{li2012chaotic}
D.~Li, M.~Han, and J.~Wang, ``Chaotic time series prediction based on a novel
  robust {Echo State Network},'' \emph{IEEE Transactions on Neural Networks and
  Learning Systems}, vol.~23, no.~5, pp. 787--799, 2012.

\bibitem{deihimi2012application}
A.~Deihimi and H.~Showkati, ``Application of {Echo State Networks} in
  short-term electric load forecasting,'' \emph{Energy}, vol.~39, no.~1, pp.
  327--340, 2012.

\bibitem{ploger2003echo}
P.~G. Pl{\"o}ger, A.~Arghir, T.~G{\"u}nther, and R.~Hosseiny, ``{Echo State
  Networks} for mobile robot modeling and control,'' in \emph{Robot Soccer
  World Cup}.\hskip 1em plus 0.5em minus 0.4em\relax Springer, 2003, pp.
  157--168.

\bibitem{ishu2004identification}
K.~Ishu, T.~van Der~Zant, V.~Becanovic, and P.~Ploger, ``Identification of
  motion with {Echo State Network},'' in \emph{MTS/IEEE Techno-Ocean}, vol.~3,
  2004, pp. 1205--1210.

\bibitem{hartland2007using}
C.~Hartland and N.~Bredeche, ``Using {Echo State Networks} for robot navigation
  behavior acquisition,'' in \emph{IEEE International Conference on Robotics
  and Biomimetics}, 2007, pp. 201--206.

\bibitem{manneschi2021SPARCE}
L.~Manneschi, A.~C. Lin, and E.~Vasilaki, ``{SpaRCe: Improved Learning of
  Reservoir Computing Systems through Sparse Representations},'' \emph{IEEE
  Transactions on Neural Networks and Learning Systems}, 2021.

\bibitem{manneschi2021exploiting}
L.~Manneschi, M.~O. Ellis, G.~Gigante, A.~C. Lin, P.~Del~Giudice, and
  E.~Vasilaki, ``{Exploiting Multiple Timescales In Hierarchical Echo State
  Networks},'' \emph{Frontiers in Applied Mathematics and Statistics}, 2021.

\bibitem{lukovsevivcius2012practical}
M.~Luko{\v{s}}evi{\v{c}}ius, ``A practical guide to applying echo state
  networks,'' in \emph{Neural networks: Tricks of the trade}.\hskip 1em plus
  0.5em minus 0.4em\relax Springer, 2012, pp. 659--686.

\bibitem{zaffar2021vpr}
M.~Zaffar, S.~Garg, M.~Milford, J.~Kooij, D.~Flynn, K.~McDonald-Maier, and
  S.~Ehsan, ``{VPR-Bench: Open-Source Visual Place Recognition Evaluation
  Framework with Quantifiable Viewpoint and Appearance Change},''
  \emph{International Journal of Computer Vision}, pp. 1--39, 2021.

\bibitem{garg2021your}
S.~Garg, T.~Fischer, and M.~Milford, ``Where is your place, visual place
  recognition?'' \emph{arXiv preprint arXiv:2103.06443}, 2021.

\bibitem{toft2020long}
C.~Toft, W.~Maddern, A.~Torii, L.~Hammarstrand, E.~Stenborg, D.~Safari,
  M.~Okutomi, M.~Pollefeys, J.~Sivic, T.~Pajdla, \emph{et~al.}, ``Long-term
  visual localization revisited,'' \emph{IEEE Transactions on Pattern Analysis
  and Machine Intelligence}, 2020.

\bibitem{glover2014day}
\BIBentryALTinterwordspacing
A.~Glover, ``Day and night, left and right,'' 2014. [Online]. Available:
  \url{https://doi.org/10.5281/zenodo.4590133}
\BIBentrySTDinterwordspacing

\bibitem{zaffar2020memorable}
M.~Zaffar, S.~Ehsan, M.~Milford, and K.~D. McDonald-Maier, ``Memorable maps: A
  framework for re-defining places in visual place recognition,'' \emph{IEEE
  Transactions on Intelligent Transportation Systems}, 2020.

\bibitem{chen2018learning}
Z.~Chen, L.~Liu, I.~Sa, Z.~Ge, and M.~Chli, ``Learning context flexible
  attention model for long-term visual place recognition,'' \emph{IEEE Robotics
  and Automation Letters}, vol.~3, no.~4, pp. 4015--4022, 2018.

\bibitem{maddern20171}
W.~Maddern, G.~Pascoe, C.~Linegar, and P.~Newman, ``1 year, 1000 km: The oxford
  robotcar dataset,'' \emph{The International Journal of Robotics Research},
  vol.~36, no.~1, pp. 3--15, 2017.

\bibitem{menzel2001cognitive}
R.~Menzel and M.~Giurfa, ``Cognitive architecture of a mini-brain: the
  honeybee,'' \emph{Trends in Cognitive Sciences}, vol.~5, no.~2, pp. 62--71,
  2001.

\bibitem{cope2018abstract}
A.~J. Cope, E.~Vasilaki, D.~Minors, C.~Sabo, J.~A. Marshall, and A.~B. Barron,
  ``Abstract concept learning in a simple neural network inspired by the insect
  brain,'' \emph{PLoS Computational Biology}, vol.~14, no.~9, p. e1006435,
  2018.

\bibitem{fahrbach2006structure}
S.~E. Fahrbach, ``Structure of the mushroom bodies of the insect brain,''
  \emph{Annual Review of Entomology}, vol.~51, pp. 209--232, 2006.

\bibitem{gerber2004engram}
B.~Gerber, H.~Tanimoto, and M.~Heisenberg, ``An engram found? evaluating the
  evidence from fruit flies,'' \emph{Current Opinion in Neurobiology}, vol.~14,
  no.~6, pp. 737--744, 2004.

\bibitem{zheng2018complete}
Z.~Zheng, J.~S. Lauritzen, E.~Perlman, C.~G. Robinson, M.~Nichols, D.~Milkie,
  O.~Torrens, J.~Price, C.~B. Fisher, N.~Sharifi, \emph{et~al.}, ``A complete
  electron microscopy volume of the brain of adult drosophila melanogaster,''
  \emph{Cell}, vol. 174, no.~3, pp. 730--743, 2018.

\bibitem{cognigni2018right}
P.~Cognigni, J.~Felsenberg, and S.~Waddell, ``Do the right thing: neural
  network mechanisms of memory formation, expression and update in
  drosophila,'' \emph{Current Opinion in Neurobiology}, vol.~49, pp. 51--58,
  2018.

\bibitem{ozdemir2021echovpr}
A.~Ozdemir, M.~Scerri, A.~B. Barron, A.~Philippides, M.~Mangan, E.~Vasilaki,
  and L.~Manneschi, ``{EchoVPR: Echo State Networks for Visual Place
  Recognition},'' \emph{arXiv preprint arXiv:2110.05572}, 2021.

\bibitem{hausler2019multi}
S.~Hausler, A.~Jacobson, and M.~Milford, ``Multi-process fusion: Visual place
  recognition using multiple image processing methods,'' \emph{IEEE Robotics
  and Automation Letters}, vol.~4, no.~2, pp. 1924--1931, 2019.

\bibitem{fischer2020event}
T.~Fischer and M.~Milford, ``Event-based visual place recognition with
  ensembles of temporal windows,'' \emph{IEEE Robotics and Automation Letters},
  vol.~5, no.~4, pp. 6924--6931, 2020.

\end{thebibliography}


\includepdf[pages=-,pagecommand={},width=\textwidth]{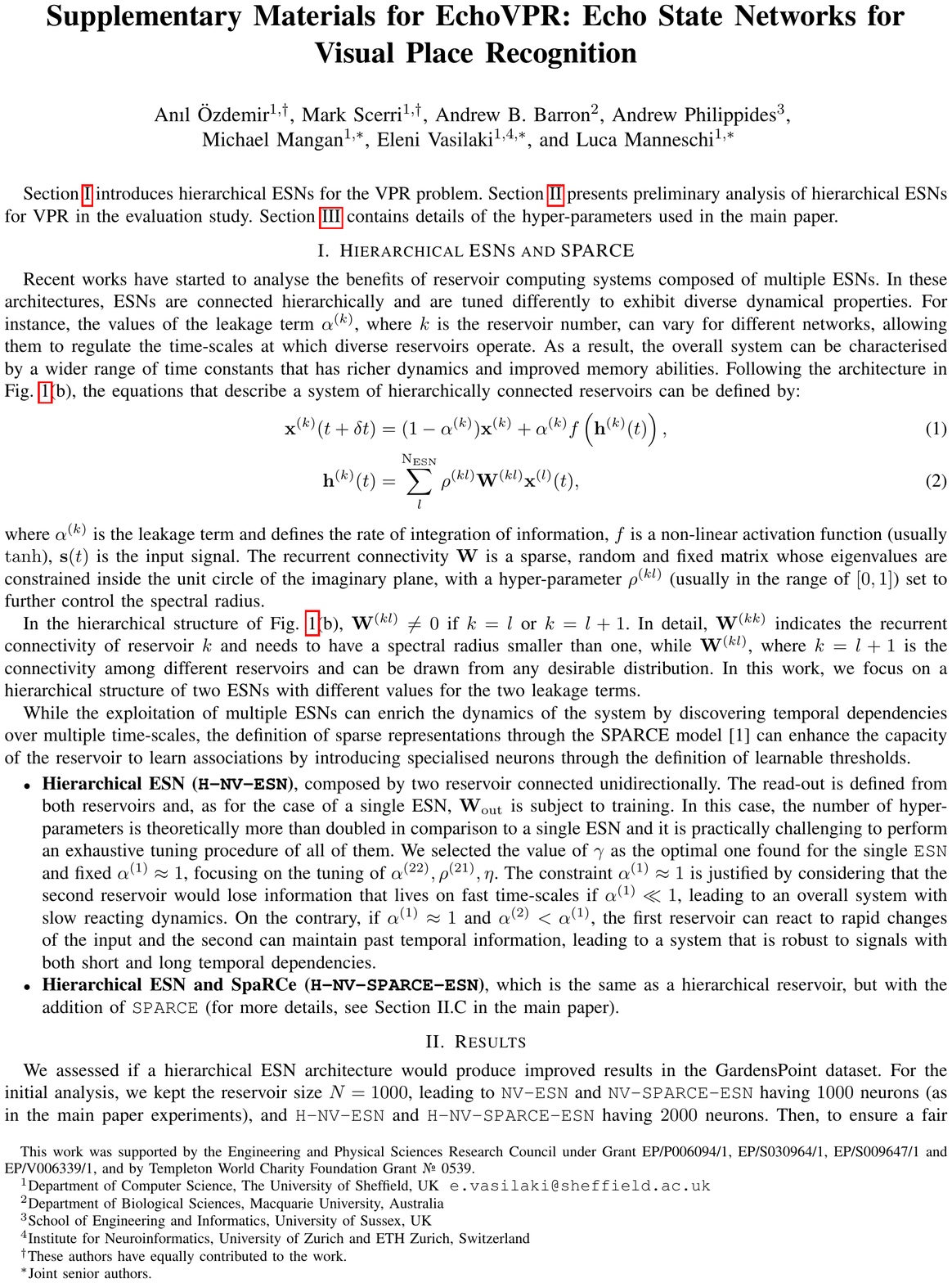}

\end{document}


\maketitle






Section~\ref{supp:hier-ESN} introduces hierarchical ESNs for the VPR problem.
Section~\ref{supp:results} presents preliminary analysis of hierarchical ESNs for VPR in the evaluation study.
Section~\ref{supp:hparams} contains details of the hyper-parameters used in the main paper. 

\section{Hierarchical ESNs and SPARCE}
\label{supp:hier-ESN}

\begin{figure*}
 \centering
    \makebox[\textwidth][c]{\includegraphics[width=0.8\textwidth]{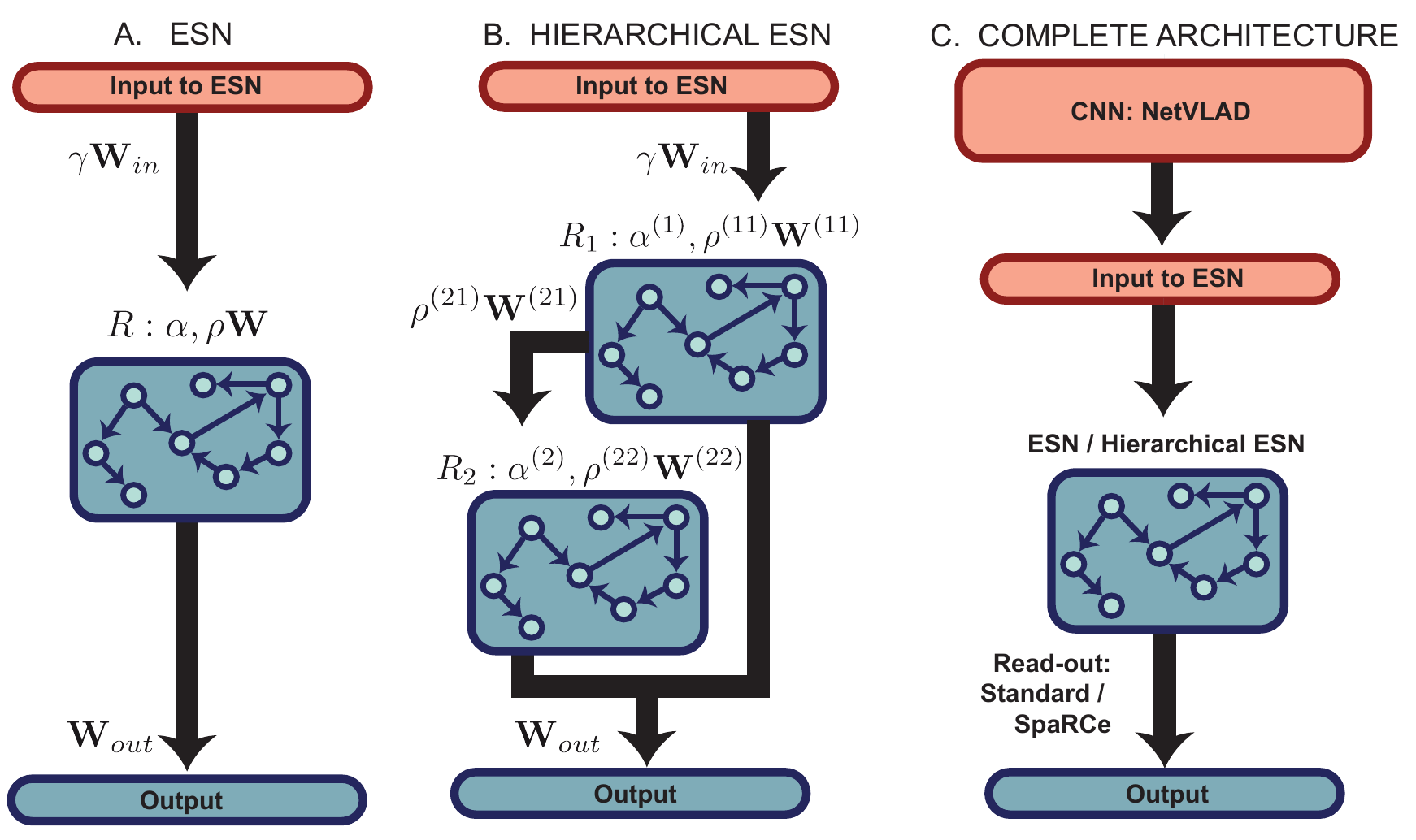}}
    \caption{
    \textbf{Scheme of the ESN models and the overall network architecture.}
    \textbf{A}: ESN protocol. 
    The input is fed to an ESN and the training process occurs on the read-out $\mathbf{W}_{\rm out}$ from the network representation. 
    When the SPARCE algorithm is adopted, additional thresholds $\mathbf{\theta}$ are initialised and adapted through the gradient.
    \textbf{B}: Hierarchical ESN. 
    The input is first processed by the first reservoir ($R_1$), which is then connected to a second ESN ($R_2$, tuned with different values of the hyper-parameters to exhibit diverse dynamical properties) unidirectionally. 
    As in $\mathbf{A}$, learning occurs on the output weights $\mathbf{W}_{\rm out}$ defined from the representation of both reservoirs and on the thresholds $\mathbf{\theta}$ when SPARCE is adopted. 
    \textbf{C}: Scheme of the overall model, composed of a pre-processing module (red boxes) and a reservoir model (blue boxes). 
    In the pre-processing, an image is fed through a CNN (i.e. NetVLAD~\cite{arandjelovic2016netvlad}), and through a hidden layer (the input to the ESN), pre-trained to reduce the dimensionality of NetVLAD output ($4096$ to $500$) and to be fed into the reservoir system. 
    The reservoir model can then be a single or hierarchical ESN with or without the SPARCE model. 
    Input images are perceived sequentially as a video, and the network has to correctly classify the location of the current image 
  }
  \label{fig:model-scheme}
\end{figure*}

Recent works have started to analyse the benefits of reservoir computing systems composed of multiple ESNs. 
In these architectures, ESNs are connected hierarchically and are tuned differently to exhibit diverse dynamical properties. 
For instance, the values of the leakage term $\alpha^{(k)}$, where $k$ is the reservoir number, can vary for different networks, allowing them to regulate the time-scales at which diverse reservoirs operate.  
As a result, the overall system can be characterised by a wider range of time constants that has richer dynamics and improved memory abilities. 
Following the architecture in Fig.~\ref{fig:model-scheme}(b), the equations that describe a system of hierarchically connected reservoirs can be defined by:
\begin{align}
    \mathbf{x}^{(k)}(t+\delta t) &=  (1-\alpha^{(k)})\mathbf{x}^{(k)} 
                                    + \alpha^{(k)} f\left( \mathbf{h}^{(k)}(t) \right), \label{eq:HESN_def}\\
    \mathbf{h}^{(k)}(t) &= \sum_{l}^{\rm N_{ESN}} \rho^{(kl)} \mathbf{W}^{(kl)}\mathbf{x}^{(l)}(t), \label{eq:HESN_def2}
\end{align}
where $\alpha^{(k)}$ is the leakage term and defines the rate of integration of information, $f$ is a non-linear activation function (usually $\operatorname{tanh}$), $\mathbf{s}(t)$ is the input signal. 
The recurrent connectivity $\mathbf{W}$ is a sparse, random and fixed matrix whose eigenvalues are constrained inside the unit circle of the imaginary plane, with a hyper-parameter $\rho^{(kl)}$ (usually in the range of $[0,1]$) set to further control the spectral radius. 

In the hierarchical structure of Fig.~\ref{fig:model-scheme}(b), $\mathbf{W}^{(kl)}\neq 0$ if $k=l$ or $k=l+1$. 
In detail, $\mathbf{W}^{(kk)}$ indicates the recurrent connectivity of reservoir $k$ and needs to have a spectral radius smaller than one, while $\mathbf{W}^{(kl)}$, where $k=l+1$ is the connectivity among different reservoirs and can be drawn from any desirable distribution.  
In this work, we focus on a hierarchical structure of two ESNs with different values for the two leakage terms. 

While the exploitation of multiple ESNs can enrich the dynamics of the system by discovering temporal dependencies over multiple time-scales, the definition of sparse representations through the SPARCE model~\cite{manneschi2021sparce} can enhance the capacity of the reservoir to learn associations by introducing specialised neurons through the definition of learnable thresholds. 

\begin{itemize}
    \item \textbf{Hierarchical ESN (\texttt{H-NV-ESN})}, composed by two reservoir connected unidirectionally. 
    The read-out is defined from both reservoirs and, as for the case of a single ESN, $\mathbf{W}_{\rm out}$ is subject to training. 
    In this case, the number of hyper-parameters is theoretically more than doubled in comparison to a single ESN and it is practically challenging to perform an exhaustive tuning procedure of all of them. 
    We selected the value of $\gamma$ as the optimal one found for the single \texttt{ESN} and fixed $\alpha^{(1)} \approx 1$, focusing on the tuning of $\alpha^{(22)}, \rho^{(21)}, \eta$. 
    The constraint $\alpha^{(1)} \approx 1$ is justified by considering that the second reservoir would lose information that lives on fast time-scales if $\alpha^{(1)} \ll 1$, leading to an overall system with slow reacting dynamics. 
    On the contrary, if $\alpha^{(1)} \approx 1$ and $\alpha^{(2)} < \alpha^{(1)}$, the first reservoir can react to rapid changes of the input and the second can maintain past temporal information, leading to a system that is robust to signals with both short and long temporal dependencies.   
    \item \textbf{Hierarchical ESN and SpaRCe (\texttt{H-NV-SPARCE-ESN})}, which is the same as a hierarchical reservoir, but with the addition of \texttt{SPARCE} (for more details, see Section II.C in the main paper). 
\end{itemize}   

\section{Results}
\label{supp:results}


We assessed if a hierarchical ESN architecture would produce improved results in the GardensPoint dataset.
For the initial analysis, we kept the reservoir size $N=1000$, leading to \texttt{NV-ESN} and \texttt{NV-SPARCE-ESN} having $1000$ neurons (as in the main paper experiments), and \texttt{H-NV-ESN} and \texttt{H-NV-SPARCE-ESN} having $2000$ neurons.
Then, to ensure a fair comparison, we increased the single reservoirs to match the size of the hierarchical models, $N=2000$.
Comparison of these three settings: (i) single reservoir models with $N=1000$, (ii) with $N=2000$ and (iii) hierarchical models with $N=2000$, is given in Table~\ref{tab:supp:hier-results}.

The results shows that the introduction of hierarchical models, compared to single reservoir $N=1000$, increased the median accuracy scores while decreasing their variance; \texttt{H-NV-ESN} median: $0.80$ and std: $0.14$ vs \texttt{H-NV-SPARCE-ESN} median: $0.87$ and std: $0.06$; both for $20$ trials). 
AUC scores showed little change, but they were already close to the maximum possible ($>0.93$), and thus there was little room for improvement. Their performance was slightly below the single reservoir $N=2000$ neurons. 
We would like to highlight that in the hierarchical models only the leakage terms ($\alpha$) and the learning rate ($\eta$) were optimised.



Considering the performance improvement consequent to the utilisation of the hierarchical model, it is evident how the GardensPoint dataset~\cite{glover2014day} contains longer temporal dependencies among images that a single ESN cannot capture. 
After an inspection of the datasets, it is clear that data of GardensPoint are captured at a higher frame-rate than the other datasets (ESSEX3IN1, SPEDTest, Corridor), where images appear more static and separated in time across each other. 
Consequently, GardensPoint has a more complex underlying temporal structure. 
At this point, we see the benefit of using hierarchical models, noting that further study is needed to assess their utilities for the VPR problem thoroughly. 

\FloatBarrier
\begin{table*}[h]
\centering
\caption{Comparison of hierarchical models against single reservoir \texttt{NV-ESN} and \texttt{NV-SPARCE}. 
$N$ is the total number of reservoir neurons.
The results are for $20$ trials.
}
\resizebox{0.65\textwidth}{!}{
\begin{tabular}{|l|c|c|}
\hline
Model                      & Accuracy mean (std) & AUC mean (std) \\
\hline
\texttt{NV-ESN} ($N=1000$)         & 0.748 (0.142)       & 0.904 (0.07)   \\
\texttt{NV-ESN} ($N=2000$)         & 0.85 (0.093)        & 0.932 (0.028)  \\
\texttt{H-NV-ESN} ($N=2000$)    & 0.829 (0.094)       & 0.913 (0.045)  \\
\hline
\texttt{NV-SPARCE-ESN} ($N=1000$)      & 0.772 (0.106)       & 0.926 (0.04)   \\
\texttt{NV-SPARCE-ESN} ($N=2000$)      & \textbf{0.867(0.059)}        & \textbf{0.951 (0.024)}  \\
\texttt{H-NV-SPARCE-ESN} ($N=2000$) & 0.858 (0.056)       & 0.936 (0.036) \\
\hline
\end{tabular}
}
\label{tab:supp:hier-results}
\end{table*}
\FloatBarrier

\section{Hyper-parameters}
\label{supp:hparams}

In this section, we provide the hyper-parameters used for the ESNs.
Table~\ref{tab:supp:hparams} shows the hyper-parameters used for datasets: GardensPoint, SPEDTest, ESSEX3IN1, Corridor, and Nordland (subset).
Table~\ref{tab:supp:hparams-longer} shows the hyper-parameter used for Nordland and Oxford RobotCar datasets.



\begin{table}[h]
    \centering
    \caption{Hyper-parameters used for experiments in Section IV. A,B,E: GardensPoint, SPEDTest, ESSEX3IN1, Corridor, and (subset) Nordland.
    For the hierarchical models, two hyper-parameters are for two reservoirs.
    }
\resizebox{0.90\textwidth}{!}{
\begin{tabular}{|l|l|l|l|l|l|}
\hline
Dataset                                         & Model                    & $\eta$ & $\alpha$     & $\gamma$       & $P_n$ \\
\hline

\multirow{6}{*}{GardensPoint}                   & \texttt{NV-ESN} (N=1000)          & 0.01          & 0.678032  & 0.000316    & $-$        \\
                                                & \texttt{NV-SPARCE-ESN} (N=1000)   & 0.01          & 0.739869  & 0.000316    & 0.4      \\
                                                & \texttt{NV-ESN} (N=2000)          & 0.0055        & 0.707946  & 0.000888    & $-$        \\
                                                & \texttt{NV-SPARCE-ESN} (N=2000)   & 0.005         & 0.723563  & 0.000888    & 0.5      \\
                                                & \texttt{H-NV-ESN} (N=2000)         & 0.001         & 0.6 / 0.9 & 0.01 / 0.01 & $-$        \\
                                                & \texttt{H-NV-SPARCE-ESN} (N=2000) & 0.0005        & 0.6 / 0.9 & 0.01 / 0.01 & 0.4      \\
\hline
\multirow{2}{*}{SPEDTest}                       & \texttt{NV-ESN}                   & 0.01          & 0.957745  & 0.00072     & $-$        \\
                                                & \texttt{NV-SPARCE-ESN}            & 0.01          & 0.978873  & 0.00072     & 0.1      \\
\hline
\multirow{2}{*}{ESSEX3IN1}                      & \texttt{NV-ESN}                   & 0.01          & 1         & 0.003728    & $-$        \\
                                                & \texttt{NV-SPARCE-ESN}            & $1 \times 10^{-5}$      & 1         & 0.003728    & 0        \\
\hline
\multirow{2}{*}{Corridor}                       & \texttt{NV-ESN}                   & 0.001         & 0.841395  & 0.003728    & $-$        \\
                                                & \texttt{NV-SPARCE-ESN}            & 0.005         & 0.841395  & 0.003728    & 0.4      \\

\hline
\multirow{2}{*}{Nordland (subset, 1000 images)} & \texttt{NV-ESN}                   & 0.01          & 1         & 0.01        & $-$        \\
                                                & \texttt{NV-SPARCE-ESN}            & 0.0001        & 1         & 0.01        & 0.25  \\  
\hline
\end{tabular}
}
\label{tab:supp:hparams}
\end{table}


\FloatBarrier
\begin{table}[!tbhp]
    \centering
    \caption{Hyper-parameters used for experiments in Section IV. C,D,E: Nordland and Oxford RobotCar.}
\resizebox{0.60\textwidth}{!}{
\begin{tabular}{|l|ll|ll|}
\hline
 & \multicolumn{2}{c|}{Nordland} & \multicolumn{2}{c|}{Oxford} \\
 & \multicolumn{2}{c|}{\scriptsize{Summer vs Winter}} & \multicolumn{2}{c|}{\scriptsize{Day vs Night}} \\
Parameter & \texttt{NV-ESN} & \texttt{NV-SPARCE-ESN} & \texttt{NV-ESN} & \texttt{NV-SPARCE-ESN} \\
\hline
 $N$ & $8000$ & $8000$ & $6000$ & $6000$ \\
 $\alpha$ & $1.0$ & $1.0$ & $0.3$ & $0.3$ \\
 $\gamma$ & $0.01$ & $0.01$ & $0.008$ & $0.008$ \\
 $\rho$ & $0.99$ & $0.99$ & $0.99$ & $0.99$ \\
 $P_n$ & $-$ & $0.0$ & $-$ & $0.7$ \\
\hline
 $N_{batch}$ & $200$ & $200$ & $30$ & $30$ \\
 $\eta$ & $1 \times 10^{-4}$ & $5 \times 10^{-4}$ & $5 \times 10^{-4}$ & $1 \times 10^{-3}$ \\
 $\eta_{\theta}$ & $-$ & $5 \times 10^{-7}$ & $-$ & $1 \times 10^{-6}$ \\
\hline
\end{tabular}
}
\label{tab:supp:hparams-longer}
\end{table}
\FloatBarrier


\bibliographystyle{IEEEtran}
\bibliography{references}